\documentclass[letterpaper]{article}
\usepackage{iccc}

\usepackage{times}
\usepackage{helvet}
\usepackage{courier}
\usepackage{balance} 
\usepackage{acronym}
\usepackage{csquotes}
\usepackage{enumerate}
\usepackage[shortlabels]{enumitem}
\usepackage{booktabs, colortbl, subcaption}
\usepackage{tikz}
\usetikzlibrary{shapes,arrows,positioning,calc,trees,chains,shapes.geometric,decorations.pathreplacing,decorations.pathmorphing,matrix,shapes.symbols,tikzmark}
\usepackage{soul}
\usepackage{transparent}
\usepackage{amsmath,amsfonts,amsthm, mathtools, yhmath}
\usepackage[switch]{lineno}
\usepackage{natbib} %
\usepackage{url}
\usepackage{mathrsfs}
\usepackage{bm}
\usepackage{balance}
\newacro{AI}{Artificial Intelligence}
\newacro{CC}{Computational Creativity}
\newacro{ECS}{exploratory creative system}
\newacro{MP}{Markov Process}
\newacroplural{MP}[MPs]{Markov Processes}
\newacro{MDP}{Markov Decision Process}
\newacroplural{MDP}[MDPs]{Markov Decision Processes}
\newacro{CSF}{Creative Systems Framework}
\newacro{POMDP}{Partially Observable Markov decision Process}
\newacroplural{POMDP}[POMDPs]{Partially Observable Markov Decision Processes}

\usepackage{xcolor}
\definecolor{bostonred}{RGB}{204,0,0}

\newcommand{\rformat}[1]{\mathcal{#1}}
\newcommand{\U}{\rformat{U}}
\renewcommand{\P}{\rformat{P}}
\newcommand{\N}{\rformat{N}}
\newcommand{\Q}{\rformat{Q}}
\newcommand{\V}{\rformat{V}}
\newcommand{\C}{\rformat{C}}
\newcommand{\B}{\rformat{B}}
\newcommand{\M}{\rformat{M}}
\newcommand{\E}{\rformat{E}}

\DeclareMathAlphabet{\mathmybb}{U}{bbold}{m}{n}

\let\oldequation\equation
\let\oldendequation\endequation

\newtheorem{definition}{Definition}%

\renewenvironment{equation}
  {\linenomathNonumbers\oldequation}
  {\oldendequation\endlinenomath}

\title{Creativity and Markov Decision Processes}%

\author{Joonas Lahikainen\textsuperscript{1} \and Nadia M. Ady\textsuperscript{1,2} \and Christian Guckelsberger\textsuperscript{1,3}\\
\textsuperscript{1}Department of Computer Science, Aalto University, Espoo, Finland\\
\textsuperscript{2}Helsinki Institute for Information Technology, Helsinki, Finland\\
\textsuperscript{3}School of Electronic Engineering and Computer Science, Queen Mary University of London, London, UK\\
christian.guckelsberger@aalto.fi
}

\begin{document}

\maketitle
\begin{abstract}
\begin{quote}
Creativity is already regularly attributed to AI systems outside specialised computational creativity (CC) communities. However, the evaluation of creativity in AI at large typically lacks grounding in creativity theory, which can promote inappropriate attributions and limit the analysis of creative behaviour. While CC researchers have translated psychological theory into formal models, the value of these models is limited by a gap to common AI frameworks. To mitigate this limitation, we identify formal mappings between Boden's process theory of creativity and Markov Decision Processes (MDPs), using the Creative Systems Framework as a stepping stone. We study three out of eleven mappings in detail to understand which types of creative processes, opportunities for (aberrations), and threats to creativity (uninspiration) could be observed in an MDP. We conclude by discussing quality criteria for the selection of such mappings for future work and applications. %
\end{quote}
\end{abstract}

\section{Introduction}

Since the inception of \ac{AI}, researchers have sought to model creativity in artificial systems \citep{mccarthy2006proposal,boden2015computational}.~%
Until recently, most \ac{CC} research has been driven by relatively small subfields of the larger AI community \citep{cook2018neighbouring}. But now, \ac{AI} at large has progressed greatly, and
creativity is now attributed to systems developed outside \ac{CC} subfields, e.g., \textit{AlphaGo}~\citep{bory2019deep, natale2022lovelace}, with increasing frequency.

Echoing, e.g., \citet{besold2016unnoticed}, we hold that many instances of \ac{AI} already exhibit creativity to an extent. 
To what extent though is an open question, as attributing creativity to such systems is usually based on intuitive judgement and not theoretically grounded.\footnote{Intuitive judgement remains a valuable type of creativity evaluation \citep{colton2012computational,natale2022lovelace}, but we contend that structured and theoretically grounded accounts are at least equally important for supporting scientific progress.} %
This may promote incorrect attributions of creativity, including failure to recognise creativity in a system entirely. 
Additionally, it limits our ability to distinguish distinct types of creativity exhibited by systems, which may inhibit system development for specific purposes. 

The required systematic reflection on the creativity of \ac{AI} can be supported by insights into how a system's components and dynamics, \textit{captured formally}, relate to creativity theory. 
Translational research between Psychology and AI has recently gained more traction \citep[e.g.,][]{van2023reclaiming, lintunen2024advancing,ady2022five}. It is notoriously hard though, in part due to the challenge of interpreting informal theory formally. %
Drawing on psychological theory, \ac{CC} researchers have developed a few formal frameworks for evaluating systems'  (potential) creativity (e.g., FACE, \citeauthor{colton2011computational}, \citeyear{colton2011computational}; Dev-ER, \citeauthor{aguilar2015dev}, \citeyear{aguilar2015dev}; CSF, \citealp{wiggins:cf_2019}). 
However, most AI systems are not originally formalised through creativity frameworks.
Assessing a given system's creativity through theory thus involves considerable interpretation, which requires effort and expertise and may introduce inconsistency.
Specifically, AI systems that engage in sequential decision-making are most prevalently formalised with \acp{MDP} \citep[p.~273]{sutton1997significance}.
A formal mapping between a creativity framework and agents in interaction with \acp{MDP} would immediately allow for the standardised analysis of numerous \ac{AI} systems in terms of how they might be considered creative. To fulfil this purpose, mapping would need to express how each mathematical object involved in one formal framework could be understood as one of the mathematical objects in the other.

Here, we propose formal mappings between AI systems modelled using \acp{MDP} %
to Margaret \citeauthor{boden:creative_mind}'s (\citeyear{boden:creative_mind}) process theory of creativity. 
We chose \citeauthor{boden:creative_mind}'s theory because it allows to distinguish different types of, obstacles to, and opportunities for creativity, while not assuming specific cognitive faculties. 
Moreover, it %
has already been formalised--and, enabled by formal rigour, further differentiated--twice in the \aclu{CSF} \citep[\acs{CSF}; ][]{wiggins:cf_2006,wiggins:comp_creativity,wiggins:cf_2019, ritchie:csf}, allowing us to use the \ac{CSF} as a stepping stone for mapping \citeauthor{boden:creative_mind}'s theory to \acp{MDP}. %

\citeauthor{colin:hierarchical_mdp_csf} \citeyearpar{colin:hierarchical_mdp_csf} made a pioneering effort to map between the \ac{CSF} and hierarchical \acp{MDP} (see Related Work). We here extend their work by proposing mappings to the more general framework of \acp{MDP}, and additionally illustrate ambiguity in our possible mappings. 
Our contributions are: (1) We argue for \acp{MDP} as a minimal framework from which to map \citeauthor{boden:creative_mind}'s theory (2) We identify eleven potential mappings between \acp{MDP} and the \ac{CSF} and evaluate three in detail; (3) We discuss which types of creative processes, uninspiration, and aberration could be observed in an \ac{MDP} under the three mappings detailed; (4) We propose quality criteria to surface mapping issues and to support selecting from mapping candidates for future analytical work and applications.
We thus follow the call made by \citet{ady2023interdisciplinary} for \ac{CC} researchers to explain choices made in selecting and interpreting definitions.

This paper thus does not conclude, but opens up an interdisciplinary research agenda, establishing a new formal basis to strengthen the dialogue between \ac{CC} and Psychology with \ac{AI} research at large. As such, it benefits and is written for an audience of multiple stakeholders. \textit{\ac{CC} researchers} receive the means to better analyse \ac{AI} systems beyond \ac{CC} to further the field's engineering and cognitive research continuum \citep{perez2018continuum}. \textit{\ac{AI} researchers} from other fields can better understand the (potential) creativity of their systems, and means to foster it. \citeauthor{boden:creative_mind}'s original theory and the \ac{CSF} have been widely used for \ac{CC} evaluation and system design, and extended in several ways \citep[e.g.,][]{wiggins:hybrid_systems, linkola:creative_societies, linkola2020action, kantosalo:co_creation_formalisation}. Our mappings to \acp{MDP} %
promise to make this \ac{CC} research heritage more relevant and accessible to AI at large. Finally, our %
mappings can allow \textit{psychologists} to simulate process theories of creativity in more diverse systems towards alleviating the formalisation and replication crises in Psychology \citep{oberauer2019addressing}. 

\section{Background}

\subsection{Boden's Process Theory of Creativity} 

Theories of creativity have been conceived from four distinct perspectives \citep{rhodes1961analysis, jordanous2016four}: the \emph{person} or \emph{producer} as originator of the work, the \emph{process} as the steps the producer undertakes when being creative, the \emph{product} as the outcome, and the \emph{press} as the sociocultural environment which shapes our views on, and the assessment of, creativity. 

We focus on process theories, which, by treating how creative products are made, have been deemed central to understanding and supporting the evaluation of creativity in natural \citep[e.g.,][]{walia2019dynamic} and computational domains \citep[e.g.,][]{colton2008creativity}, and have been promoted for including \enquote{many important ideas that can and
should influence the design of a creative system}~\citep{lamb2018evaluating}. 

Crucially though, from \citeauthor{wallas1926art}' (\citeyear{wallas1926art}) classic four-stage model to \citeauthor{green2023process}'s (\citeyear{green2023process}) recent definition, many theories of the creative process assume human cognitive features such as attention or unconscious reasoning, limiting their applicability to artificial systems \citep{lamb2018evaluating}.
A notable exception is the theory of Margaret \citeauthor{boden:creative_mind}, who distinguishes three types of creative \textit{processes}: combinatorial, exploratory, and transformational. These rest on what \citeauthor{boden:creative_mind} denotes a \textit{conceptual space}: 
a structured way of thinking which both constrains and makes possible a particular variety of thoughts (\citeyear[p.~58]{boden:creative_mind}).
It can be conceived as a space of all complete and incomplete things (including both mental concepts and physical artefacts%
) that could be generated according to a set of (agreed) rules.

\textit{Combinatorial creativity} refers to the creation of new concepts by combining features of existing ones. \textit{Exploratory creativity} corresponds to exploring the conceptual space for new and valued concepts. Finally, \citeauthor{boden:creative_mind} introduces \textit{transformational creativity} as reaching new points only made accessible by altering the rules defining the space itself \enquote{so that thoughts are now possible which previously (...) were literally inconceivable}~\citep[][p.~6]{boden:creative_mind}. Many process theorists consider transformational creativity as the most profound \citep[summarised by][]{lamb2018evaluating},
including\citeauthor{boden2009computer} herself, who notes that transformational creativity is \enquote{the \enquote{sexiest} of the three types, because it can give rise to ideas that are not only new but fundamentally different from any that went before} \citep[p.~25]{boden2009computer}. %
For this reason, an important focus in our paper will be on how systems, as viewed through our mappings, might exhibit transformational creativity.

\subsection{The Creative Systems Framework}

Abstaining from assumptions of human cognitive facilities, \citeauthor{boden:creative_mind}'s theory has unsurprisingly become the most popular process theory of creativity in \ac{CC}~\citep{lamb2018evaluating}. \citet{wiggins:cf_2006,wiggins:comp_creativity,wiggins:cf_2019} formalised the theory in set-theoretic terms, hereby providing a more concrete interpretation and adding to \citeauthor{boden:creative_mind}'s account. \citet{ritchie:csf} later re-formulated and extended \citeauthor{wiggins:cf_2006}' framework while retaining backward compatibility. We rely on \citeauthor{ritchie:csf}'s formulation because it simplifies the formalism, dropping \citeauthor{wiggins:cf_2019}'s reliance on a universal language and interpreters. Moreover, it has been used by \citet{colin:hierarchical_mdp_csf} in the only instance of related work; using the same framework thus also eases comparison. 
The definitions that follow are taken from \citet{ritchie:csf} and \citet{wiggins:cf_2019}, modified only for brevity or clarity; most of our notation is that of \citet[see p.~43 for relation to \citeauthor{wiggins:cf_2019}]{ritchie:csf}.
\begin{definition}[General notation]\
    \begin{enumerate}[wide=0pt,leftmargin=\parindent]%
        \item For any sets $A$ and $B$, $B^A$ denotes the set of functions from $A$ to $B$. In particular, $[0, 1]^A$ denotes the set of functions from $A$ to real values between 0 and 1, inclusive.
        \item For any set $A$, $\textit{tuples}(A)$ denotes the set of all finite tuples of elements in $A$; e.g.,~if $A = \{1,2,3\}$, then $\textit{tuples}(A) = \{1,2,3,(1,1),(1,2),\dots\}$.
        \item For any set of tuples $X$, we define the set of distinct elements in the tuples of $X$ as $\textit{elements}(X) \coloneqq \{x \mid \exists (y_1,\dots,y_n) \in X \land \exists\ i\ \in \{ 1, \dots, n\} : x = y_i\}$. This flattens a set of tuples into a set of distinct elements: for set $A$ and $X = \textit{tuples}(A)$, we have $\textit{elements}(X) = A$.
        \item For any set $A$, function $f \in [0,1]^A$, and threshold $\alpha \in [0,1]$, we define the \emph{strong $\alpha$-cut of $A$} as $f^{>\alpha}(A) \coloneqq \{a \in A \mid f(a) > \alpha\}$.
    \end{enumerate}
\end{definition}

To formalise transformational creativity, \citeauthor{wiggins:cf_2019} introduced a set representing all conceivable concepts. %
\begin{definition}[Universe]
    The universe, $\U$, is a set (specifically a multidimensional space; \citealp[p.~26]{wiggins:cf_2019}) capable of representing anything, and all possible distinct concepts correspond with distinct points in $\U$.
\end{definition}

\noindent\textbf{Axiom 1} (\textit{Universality}). All possible concepts, including the empty concept, $\top$, are represented in $\U$.

\noindent\textbf{Axiom 2} (\textit{Non-identity of concepts}). All concepts represented in $\U$ are non-identical, meaning $\forall c_1, c_2 \in \U, c_1 \neq c_2.$

\noindent\textbf{Axioms 3 \& 4} (\textit{Universal inclusion}). All conceptual spaces (3) are strict subsets of $\U$ and (4) include $\top$.

\smallskip

\citet[p.~43]{ritchie:csf} adapts \citeauthor{wiggins:cf_2006}' formulation to define an \emph{exploratory creative system}%
\footnote{We adopt the name \enquote{exploratory \emph{creative} system} by convention. We do not want to convey the impression that such a system exhibits creativity at any moment. Instead, we hold that it has the \textit{potential} to exhibit creativity as defined by \citet{boden:creative_mind}.}
as a 4-tuple:

\begin{definition}[Exploratory Creative System, ECS]\acused{ECS}
    We define an \emph{\ac{ECS}} $\E$ as a 4-tuple $(\P,\N,\V,\Q)$ consisting of:
    \begin{enumerate}[wide=0pt,leftmargin=\parindent]
        \item $\P \subseteq \U$, {sub-universe}
        \item $\N \in [0, 1]^\P$, {acceptability function}
        \item $\V \in [0, 1]^\P$, {evaluation function}
        \item $\Q:[0, 1]^P\!\! \times\! [0, 1]^\P\!\! \to\! \textit{tuples}(\P)^{\textit{tuples}(\P)}$, {traversal strategy}%
    \end{enumerate}
    \label{def:exploratory_system}
\end{definition}

A \textit{conceptual space}, defined below, is the set of acceptable concepts out of a subset, $\P$, of all concepts. 
Restriction to subset $\P \subseteq \U$ (a \emph{sub-universe}) allows us to make distinctions between systems w.r.t.\ their access to the full universe \citep[p.~42]{ritchie:csf}. 

\begin{definition}[Conceptual space]
    For acceptability function $\N \in [0, 1]^\P$, acceptability threshold $\alpha \in [0,1]$, and sub-universe $\P \subseteq \U$, we define the conceptual space $\C$ as
    \label{def:conceptual_space}
    \begin{equation}
    \C \coloneqq \N^{>\alpha}(\P) = \{c \in \P \mid \N(c) > \alpha\}
    \end{equation}
\end{definition}

While, according to \citeauthor{wiggins:cf_2019}, evaluation, $\V$, does not play into the \emph{definition} of a conceptual space (see Critical Assumptions), value is a core requirement of creativity \citep[e.g.,][]{runco2012standard}. Thus, not only the acceptability function, $\N$, but also the evaluation function, $\V$, influence the traversal strategy, $\Q$.
The ECS starts at an initial concept and \enquote{searches} through the conceptual space by means of its traversal strategy. 
In the \ac{CSF}, \textit{exploratory creativity} corresponds to exploring the conceptual space $\C$ for new concepts valued in terms of $\V$. 

\citet[p.~454]{wiggins:cf_2006} offered a mechanistic explanation of transformational creativity by formalising it as exploratory creativity, but on a \emph{meta} level. A meta-level creative system, $(\P^\textit{meta},\N^\textit{meta},\V^\textit{meta},\Q^\textit{meta})$, searches a conceptual space in which the concepts are triples $(\N,\V,\Q)$, with the sub-universe, $\P^\textit{meta}$, being the set of all such triples \citep[p.~44]{ritchie:csf}. 
Recursively, a creative system could have many such levels. 
The lowest level is considered the \textit{object}-level system.

\citeauthor{boden:creative_mind} (\citeyear[p.~58]{boden:creative_mind}) originally explained {transformation} as modifying the existing rules of a conceptual space to make possible concepts that were not possible before. \citeauthor{wiggins:cf_2019} extended the notion of transformation to changing the traversal strategy, noting that such a change might \enquote{make accessible concepts which were not previously available} to the agent (\citeyear[p.~32]{wiggins:cf_2019}). We therefore consider:%
\begin{itemize}
    \item \textit{$\N$-transformation}: modifying the conceptual space via changes to the acceptability function. 
    \item \textit{$\Q$-transformation}: changing the traversal strategy. 
\end{itemize} 
For the purpose of assessing transformational creativity, 
we follow \citeauthor{wiggins:cf_2019}' (\citeyear[p.~33]{wiggins:cf_2019}) suggestion that a transformation is valued if, given a fixed object-level evaluation function $\V$, the transformation admits new concepts valued under $\V$, either to the set of reachable concepts (defined below) or the conceptual space itself. Following \citeauthor{boden:creative_mind}'s (\citeyear[p.~10]{boden:creative_mind}) requirements for creativity to involve both novelty and value, we assume a transformation can only be considered transformational creativity if it is valued.

\citet[p.~456]{wiggins:cf_2006} additionally introduced \textit{aberrations}, characterising the traversal strategy $\Q$ reaching a set of concepts $\B$ which lie outside the conceptual space \citep[cf.][p.~45]{ritchie:csf}. Aberrations could be used to trigger and guide transformations as opportunities for creativity.

\begin{definition}[Set of reachable concepts]
Let \ac{ECS} $\E = (\P, \N, \V, \Q)$. Then, starting from an initial tuple of concepts $B$, we denote the set of reachable concepts within $\E$ in $m$ steps by
\begin{equation}
    \E^m(B) \coloneqq \textit{elements}\left(\bigcup_{n=0}^m \Q(\N, \V)^n(B)\right)
\end{equation}
where exponent $n$ denotes repeated applications of traversal strategy $\Q$. Notably, $B$ can be given by $\top$, i.e. the empty concept, corresponding to a blank canvas.
\label{def:reachability}
\end{definition}

An aberration occurs when $
    \B \coloneqq \E^\infty(B) \setminus \C
$
is non-empty. Aberrations fall into three categories depending on how this set $\B$ is valued, given threshold $\beta \in [0,1]$:\footnote{Crisp sets ease discussion, but yield binary descriptions of conceptual spaces, \textit{uninspiration}, and \textit{aberration}, contradicting \citeauthor{boden:creative_mind}'s philosophy (\citeyear[p.~2]{boden:creative_mind}): \enquote{Rather than asking \enquote{Is that idea creative, yes or no?} we should ask \enquote{Just how creative is it, and in just which way(s)?}} However, modifying the \ac{CSF} is out of scope.} \textit{perfect} ($\V^{>\beta}(\B){=}\B$), \textit{productive} ($\V^{>\beta}(\B){\subset}\B$) and \textit{pointless} ($\V^{>\beta}(\B){=} \emptyset$) aberrations. 

\citeauthor{wiggins:cf_2019} furthermore characterises three cases of \textit{uninspiration}, causing the system to fail at being creative \citep[notation adopted from][pp.~43--46]{ritchie:csf}:
\begin{itemize}
\item \textit{Generative uninspiration.} When the traversal strategy does not find any valued concepts, $\V^{>\beta}(\E^\infty(\top)) = \emptyset$.
\item \textit{Conceptual uninspiration.} When there are no valued concepts in the conceptual space, $\V^{>\beta}(\C) = \emptyset$.
\item \textit{Hopeless uninspiration.} When there are no valued concepts in the sub-universe, $\V^{>\beta}(\P) = \emptyset$.\footnote{\citeauthor{ritchie:csf}'s \citeyearpar[pp.~43, 45]{ritchie:csf} re-definition of hopeless uninspiration is contradictory, leaving unclear whether it refers to a lack of valued concepts in the \textit{sub-universe}, $\P$, or \textit{universe}, $\U$.}
\end{itemize}
A threat to creativity, uninspiration complements aberration. 

\subsection{Markov Decision Processes}
A \textit{Markov Process} \citep[\textit{MP};][]{markov:dependent_events, levin:markov_chains}\acused{MP} models stochastic \textit{transitions} between \textit{states} of a system. As per the \textit{Markov property}, the probability of moving to a specific next state depends only on the current state. 
The structure of \acp{MP} often remains useful even if the Markov property does not exactly hold, and so \acp{MP} are used abundantly across many academic fields to model the dynamics of natural and artificial systems, and especially within \ac{AI}.

A \textit{\acf{MDP}} extends \acp{MP} by introducing a notion of agency in the form of \textit{actions} that influence the transition between states. Actions are chosen based on the current state via a \textit{policy}. Moreover, \acp{MDP} define \textit{rewards} as scalar feedback from state transitions to %
define the problem to be solved with a suitable policy.

\begin{definition}[Markov Decision Process]
    \label{def:mdp}
    We define a discrete-time \ac{MDP} as a 4-tuple $(S,A,T,R)$, consisting of %
    state space $S$, %
    action space $A$, stationary transition probabilities $T$, and stochastic reward function $R$.
\end{definition}
In this paper, we chose to focus on discrete-time \acp{MDP} for its fit with the use of discrete \enquote{steps} in the \ac{CSF}.
We denote the probability of transitioning from state $s$ to the next state $s'$ with action $a$ as
\begin{equation}
    T_{a}(s,s') \coloneqq P(s'\mid s,a)
\end{equation}
with the intuition that each action results in a unique transition matrix indexed by states. 
We assume both the reward function $R: S \times A \times S \to \mathbb{R}$ and policy $\pi: S \times A \to [0,1]$ to be stochastic, without loss of generality. 
While the stochastic policy is characterised by a probability distribution, our stochastic reward function directly samples rewards from an underlying reward distribution. For generality, the output of $R(s,a,s')$ need only depend on a subset of these inputs. Excluding the policy from Definition \ref{def:mdp} lets us discuss transitioning through the same \ac{MDP} with different policies.

The goal for the agent is to maximize its \textit{return}. Return, $G_t$, can be formalized in multiple ways, but is a function of the rewards observed after time $t$: $R(s_t,a_t,s_{t+1}), R(s_{t+1},a_{t+1},s_{t+2}), \cdots$. Different formulations of return include discounted or undiscounted sums of rewards or as the sum of differences between received rewards and the average reward \cite[pp.~54--55,~249--250]{sutton:rl_intro}; we do not assume a particular formulation here. Under a given policy, $\pi$, the return allows us to define the \emph{value} of each state (\textit{ibid}, p.~54): \begin{align}v_\pi(s) := \mathbb{E}_\pi \left[ G_t \middle| s_t = s\right]\end{align}
The definition of return changes the problem the agent is set to solve, but in each case, solving the problem means finding an optimal policy $\pi^*$ such that $v_{\pi^*}(s) \geq v_{\pi}(s)$ for all states $s\in S$ and all policies $\pi$ (\textit{ibid}, p.~62).

 Such an optimal policy $\pi^*$ as one solution to the sequential decision-making problem defined by the MDP can be found with a plethora of techniques, e.g.,~from reinforcement learning \citep{sutton:rl_intro}. Some techniques maintain an estimate of some $v_{\pi}(s)$, which we denote by $V$.

\acp{MDP} are part of a 
family of extensions to \acp{MP}, 
including \acp{POMDP}, which extend \acp{MDP} with partial observations of the state, and Hierarchical \acp{MDP}, which extend \acp{MDP} with temporally extended actions and a hierarchical structure. %
We focus here on \acp{MDP} for their generality and widespread application \citep[p.~273]{sutton1997significance}, but other members of this larger family may be worth considering for other mappings \citep[cf.][Related Work]{colin:hierarchical_mdp_csf}.

\section{Mappings}
In this paper, we seek to construct mappings from agents in interaction with \acp{MDP} to the \ac{CSF}. \Ac{AI} systems employ numerous different algorithms and architectures, but with \acp{MDP}, one way of abstracting at least part of their agency\footnote{Defining agency in \ac{AI} systems is an ongoing effort, cf.~e.g.,~\citet{biehl2022interpreting, kenton2023discovering}.} or individuality is via their policy. For this reason, we model an agent in interaction with an \ac{MDP} as a pairing between an \ac{MDP} and a (potentially non-stationary) policy. \textbf{Our goal, then, is to understand any given \ac{MDP}-policy pair as an \ac{ECS}.} %
We thus aim to construct mappings of the form: 
\begin{align}
\M: &\left\{ (S,A,T,R, \pi) \middle| \begin{array}{r} (S,A,T,R) \text{ an \ac{MDP}, }\\ \pi \text{ a policy}\end{array} \right\} \label{eq:mapform} \\ 
&\to \left\{ (\U, \P, \N, \V, \Q) \middle| \begin{array}{r} \U \text{ a universe}, \\(\P \subseteq \U, \N, \V, \Q) \text{ an \ac{ECS}}\end{array}\right\} \nonumber
\end{align}
Ideally, each mapping should be a total function, capable of mapping any \ac{MDP}-policy pair to an \ac{ECS}. We can then use the \ac{CSF} to analyse any system that is modelled as a policy over an \ac{MDP}. 
By virtue of this choice, we can %
retain the expressivity of \citeauthor{boden:creative_mind}'s framework. This in turn enables studying the extent to which different types of creative processes, opportunities for, and obstacles to creativity are expressed in specific \ac{MDP} instances, or the \ac{MDP} framework in general.

We map agents in interaction with \acp{MDP} to \citeauthor{boden:creative_mind}'s theory, but usually only capture agents w.r.t. their policy; In some cases, though, we also used the agent's estimated value function (extending the 5-tuple in the domain shown in Equation \ref{eq:mapform} to a 6-tuple to include $V$). We were initially concerned that this choice would be limiting, as many algorithms do not use estimated value functions \citep[see, e.g.,][p.~321]{sutton:rl_intro}. However, we only use $V$ in defining the evaluation function $\V$. Since $\Q$ only can, but does not have to be, dependent on $\V$ \citep[p.~29]{wiggins:cf_2019}, with careful choice of $\Q$, $V$ can be safely ignored if it is not implemented in a system of interest. This allows the mapping to retain the semantic similarity of the value function to \emph{evaluation}. Below, we explore in one mapping the use of rewards (which \enquote{determine the immediate, intrinsic desirability of environmental states}) and in two others the use of values (which \enquote{indicate the long-term desirability of states}) for their semantic similarity to the \textit{value} component of evaluation \citep[p.~6]{sutton:rl_intro}.

Under the paradigm of sequential decision-making and creative autonomy \citep{jennings2010developing, saunders2012towards}, we assume exploration, transformation, abberation, and uninspiration to be driven by the agent, rather than a separate cause (such as the system designer). Therefore, in our interpretations of these dynamics, we only consider elements that the agent can change, notably excluding $S, A, T$, and $R$.  

\subsection{\acp{MDP} as Mapping Domain} 
\acp{MDP} are very widely used \citep[Section~1.7]{sutton:rl_intro}, but are an extension of a simpler process model: \acp{MP}. 
If we mapped \acp{MP}, we could map any \ac{MDP} by extension. However, we determined that mapping \acp{MP} is not appropriate since the \ac{CSF} requires a means to \textit{evaluate} concepts; defining an \ac{ECS} requires a choice of evaluation function $\V$. Some authors consider evaluation to be a minimal requirement for creative autonomy \citep{jennings2010developing} and transformational creativity \citep{wiggins:cf_2019}. Neither states nor transition probabilities (nor combination of them) have similar meaning to the evaluation function in the \ac{CSF}.

Intuitively, \acp{MDP} as extensions of \acp{MP} appear as the most natural next mapping domain in that the evaluation function in the \ac{CSF} has clear analogues in \ac{MDP} defined in terms of the reward function. We recognise that further generalisations of \acp{MDP} such as \acp{POMDP} and Hierarchical \acp{MDP} would offer interesting viewpoints for mappings; however, they are in less widespread use than \acp{MDP} and their use would further expand the space of potential mappings, resulting in more candidate mappings to select from. Further benefits over \acp{MDP} in terms of semantic or formal connections to \citeauthor{boden:creative_mind}'s theory, as well as \citeauthor{wiggins:cf_2019}' and \citeauthor{ritchie:csf}'s \ac{CSF} are not apparent.

\subsection{Mapping Procedure}
Within the limited scope of this paper, we only discuss a selection of mappings without being exhaustive. 
Mirroring two-phase models of creativity \citep[e.g.,][]{kleinmintz2019two}, our mapping procedure involved two phases: a generation phase followed by an evaluation phase. In the generation phase, we explored potential mapping candidates, beginning with the decision of what aspect of the \ac{MDP} and agent might be mapped to concepts (elements of $\U$).  
Table~\ref{tbl:conceptual_spaces} lists all candidates we considered. In this phase, we considered not only the components listed on the left-hand side of Equation \ref{eq:mapform}, but also policy-learning algorithms and their hyperparameters. 

\begin{table}[htb]
    \centering
    \begin{tabular}{ll}
        \toprule
        Mapping & Concepts $c \in \C$\\
        \midrule

        $\M_l$ & Policy-learning algorithms $l \in L$\\
        
        $\M_{\lambda}$ & Hyperparameters $\lambda_l \in \mathbb{R}^n$\\

        $\M_{l,\lambda}$ & $(l,\lambda_l) \in L \times \mathbb{R}^n$\\

        $\M_\pi$ & Policies $\pi \in \Pi$\\

        $^\star\M_s$ & States $s \in S$\\

        $\M_{a}$ & Actions $a \in A$\\

        $\M_{r}$ & Rewards $r \in \mathbb{R}$\\
        
        $\M_{s,a}$ & Tuples $(s,a) \in S \times A$\\

        $\M_{s,a,r}$ & Tuples $(s,a,r) \in S \times A \times \mathbb{R}$\\

        $^\star\M_{s,a,s'}$ & Transitions $(s,a,s') \in S \times A \times S$\\

        $^\star\M_\tau$ & Trajectories $(s,a,s',\dots,s_\textit{last}) \in Tr$\\ %
        \bottomrule
    \end{tabular}
    \caption{Possible conceptual spaces identified in the exploratory stage. Starred $^\star$ options were examined further.}
    \label{tbl:conceptual_spaces}
\end{table}

\begin{table*}[t]
    \small
    \centering
    \begin{tabular}{llll}
        \toprule
        \ac{CSF} & $\M_{s}$ & $\M_{s,a,s'}$ & $\M_\tau$\\
        \midrule
        
        Concept $c$ & State $s$ & Transition $\delta = (s,a,s')$ & Trajectory $\tau = (s,a,s',\dots,s_\text{last})$\\
        
        \textbf{First-order definitions}\\
        Universe $\U$ & $\bigcup_S S$ & $\bigcup_{S,A}S\! \times\! A\! \times\! S$ & $\bigcup_{S,A}\textit{tuples}(S\! \times\! A)$\\

        Sub-universe $\P \subseteq \U$ & $\bigcup_S S$ & $S\! \times\! A\! \times\! S$ & $\textit{tuples}(S\! \times\! A)$\\

        Acceptability $\N$ & Membership function $\mu$ of $S$ & $p_{\delta}(\delta) := T_a(s,s')\pi(\mathrel{a}\mid s)$ & $p_{\tau}(\tau) := P(\mathrel{\tau}\mid s,\pi,T)$\\
        
        Evaluation $\V$ & $\hat{V} := \text{ normalised }V$ & $\hat{R} :=\text{ normalised }R$ & $\overline{V} :=\text{ normalised } V(s_\text{last})$\\ %

        Traversal strategy $\Q(\N,\V)$ & $f: \textit{tuples}(S) \to \textit{tuples}(S)$ & $f: \textit{tuples}(\P) \to \textit{tuples}(\P)$ & $f: \textit{tuples}(\P) \to \textit{tuples}(\P)$\\

        \multicolumn{2}{l}{}\\
        \textbf{Higher-order definitions}\\
        Conceptual space $\C$ & $\mu^{>\alpha}(\P)$ & $p_{\delta}^{>\alpha}(\P)$ & $p_{\tau}^{>\alpha}(\P)$\\
        $\N$-transformation & — & $\pi \to \pi'$, s.t. $\C \neq \C'$ & $\pi \to \pi'$, s.t. $\C \neq \C'$\\
        $\Q$-transformation & $\pi \to \pi'$ & $\pi \to \pi'$ & $\pi \to \pi'$\\

        Aberration          & — & Reaching $\delta$s s.t. $p_{\delta}(\delta) \leq \alpha$ and & Reaching $\tau$s s.t. $p_\tau(\tau) \leq \alpha$ and\\
        \quad Perfect       & \quad— & \quad all $\delta$s are rewarding.       & \quad all $\tau$s are valued.\\
        \quad Productive    & \quad— & \quad some $\delta$s are rewarding.       & \quad some $\tau$s are valued.\\
        \quad Pointless     & \quad— & \quad no $\delta$s are rewarding.      & \quad no $\tau$s are valued.\\

        Uninspiration       & No $s \in \hat{V}^{>\beta}(\P)$         & No $\delta \in \hat{R}^{>\beta}(\P)$       & No $\tau \in \overline{V}^{>\beta}(\P)$\\
        \quad Generative    & \quad can be found with $\pi$.  & \quad can be found with $\pi$.          & \quad can be found with $\pi$.\\
        \quad Conceptual    & \quad exists in $S$.               & \quad has $p_\delta(\delta) > \alpha$. & \quad has $p_\tau(\tau) > \alpha$.\\
        \quad Hopeless      & \quad exists in $\bigcup_S S$     & \quad exists in $\P$.     & \quad exists in $\P$.\\

    \end{tabular}
    \caption{Mappings $\M_s, \M_{s,a,s'}$ and $\M_\tau$. Notation: apostrophe $'$ indicates succession (state $s'$ follows state $s$), \enquote{—} denotes no mapping, the union of all conceivable sets (excluding the union itself) is denoted as $\bigcup_X X$, $\mu: \P \to \{0,1\}$ is the membership function for state space $S$, $f$ is defined by outputting the result of a one-step rollout of policy $\pi$ on each concept in its input. Higher-order definitions are derived from first-order definitions.}
    \label{tbl:all_mappings}%
\end{table*}

In the evaluation phase, we filtered our initial candidates, aiming to keep a small but diverse sample to discuss in detail. 
In this phase, we followed our intuition about which candidates best match \citeauthor{boden:creative_mind}'s theory and its extension via the \ac{CSF}. Further below, we provide a detailed reflection on our intuitive choices, informing potential quality criteria for assessing mappings. Here, we only briefly express why we excluded eight of our original candidates.
We dropped mapping $\M_\pi$, for which exploratory creativity would correspond to the discovery of new and valued policies. This is essentially what \ac{MDP} solvers do, but it resembles more what \citeauthor{wiggins:cf_2019} and \citeauthor{ritchie:csf} describe as meta-level exploration, leaving ambiguity at the object-level. 
The same applies to $\M_l$, $\M_\lambda$ and $\M_{l,\lambda}$ which were consequently also excluded. 
We excluded $\M_a$ since many applications of \acp{MDP} use discrete and small action spaces, which would be quickly exhausted. We moreover excluded $\M_r$ and $\M_{s,a,r}$, as understanding a reward as representing a concept does not seem intuitive. This left us with groups of mappings from state-action tuples or trajectories to concepts, from which we picked one each for diversity. 
We chose $\M_{s,a,s'}$ over $\M_{s,a}$ because 
we saw very natural mappings for the acceptability and evaluation functions, as the transition-probability function and reward function both take $(s,a,s')$ as input.
This leaves us with the mappings $\M_s$, $\M_{s,a,s'}$ and $\M_\tau$, which we detail in Table~\ref{tbl:all_mappings} and the following subsections, and which form the basis of our discussion.

\subsection{\texorpdfstring{Mapping $\M_s$}{Ms}}

\noindent\textbf{Universe \& Sub-universe.} We map both \textit{universe} $\U$ and \emph{sub-universe} $\P$ to the union of all conceivable state spaces $\bigcup_S S$. Consequentially, concepts are mapped to states.

\smallskip
\noindent\textbf{Acceptability \& Conceptual Space.}
We map \textit{acceptability} $\N$ to a membership function $\mu: \bigcup_S S \to \{0,1\}$ which outputs one if the input state belongs to $S$ and zero otherwise. That is, this function describes the logic by which states are included in state space $S$. As a result, \textit{conceptual space} $\C$ is mapped to state space $S$, as long as the acceptability threshold $\alpha < 1$; if $\alpha = 1$, then $\C = \emptyset$ (a consequence of Def.~\ref{def:conceptual_space}). 

\smallskip
\noindent\textbf{Evaluation.}
We map \textit{evaluation} $\V$ to a normalisation $\hat{V}: \bigcup_S S \to [0,1]$ of the agent's estimated value function $V$. %

\smallskip
\noindent\textbf{Traversal Strategy.}
\textit{Traversal strategy} $\Q$ maps 
to a one-step rollout of policy $\pi$: $s' \sim T_{a \sim \pi(s)}(s, \cdot)$ for every state $s$ in its input. That is, $\Q$ takes a tuple of current states as input, and, for each state $s$, samples action $a$ from $\pi(s)$, and outputs the following state $s'$ to compose an output tuple. $\Q$ traversing from tuples of concepts to tuples of concepts is directly from the \ac{CSF} \cite[p.~35]{wiggins:cf_2019}.\footnote{Appealingly, potential to traverse multiple rollouts at once may be suited to more general agents including asynchronous \citep{mnih2016asynchronous} and model-based agents \cite[Ch.~8]{sutton:rl_intro}.}

\smallskip
\noindent\textbf{Transformations.}
\textit{$\N$-transformation} would require the conceptual space $\C$ to change, but $\C$ corresponds to the state space $S$, which we assume cannot be modified by the agent. %
\textit{$\Q$-transformation} maps to a change in policy $\pi$.

\smallskip
\noindent\textbf{Aberration \& Uninspiration.}
\textit{Aberrations} are left unmapped since finding states outside of state space $S$ is theoretically impossible. %
\textit{Generative uninspiration} here reflects policy $\pi$ performing poorly with respect to normalised value function $\hat{V}$, as none of the reachable states exceed value threshold $\beta$ when evaluated by $\hat{V}$. 
Generative uninspiration may reflect either a truly suboptimal policy or a poorly estimated value function. 
\textit{Conceptual uninspiration} suggests an ill-formed 
value function $\hat{V}$ or value threshold $\beta$ where no states in $S$ are sufficiently valued. \textit{Hopeless uninspiration} means no state in any \ac{MDP} would be sufficiently valued.

\subsection{Mapping \texorpdfstring{$\M_{s,a,s'}$}{Msas}}

\noindent\textbf{Universe \& Sub-universe.}
\textit{Universe} $\U$ comprises all conceivable state transitions in all potential \acp{MDP}, $\delta \in \bigcup_{S,A}S\! \times\! A\! \times\! S$. 
\textit{Sub-universe} $\P$ then maps to a narrowed down version $S \times A \times S$ defined by a particular $S$ and $A$ in the mapped \ac{MDP}.

\smallskip
\noindent\textbf{Acceptability \& Conceptual Space.} \textit{Acceptability} $\N$ maps to probability $p_{\delta}$, where $p_{\delta}(\delta) := T_a(s,s')\pi(a \vert s)$. \textit{Conceptual space} $\C$ then contains transitions with $p_{\delta}(\delta) > \alpha$.

\smallskip
\noindent\textbf{Evaluation.} We map \textit{evaluation} $\V$ to the normalised reward function $\hat{R}: S\! \times\! A\! \times\! S \to [0,1]$.

\smallskip
\noindent\textbf{Traversal Strategy.} \textit{Traversal strategy} $\Q$ maps transitions to the next transition triple, given a one-step rollout of policy $\pi$: that is, transition $(s,a,s')$ maps to $(s',a',s'')$ where $s'' \sim T_{a' \sim \pi({s'})}(s', \cdot)$. As in $\M_s$, inputs and outputs can be tuples.

\smallskip
\noindent\textbf{Transformations.} Both \textit{$\N$-transformation} and \textit{$\Q$-transformation} map to changes in policy $\pi$. However, while any change to $\pi$ corresponds to a $\Q$-transformation, $\N$-transformation additionally requires that the conceptual space changes. That is, the probability of some transition previously $\leq\alpha$ must now exceed it, or one previously $>\alpha$ must drop below or meet it. 

\smallskip
\noindent\textbf{Aberration \& Uninspiration} \textit{Aberration} here refers to experiencing transitions which have at most probability $\alpha$ of occurring. These aberrations are then further categorised into \textit{perfect}, \textit{productive}, and \textit{pointless}, depending on whether all, only some, or none of these aberrant transitions produce rewards exceeding value threshold $\beta$.

\textit{Uninspiration} here refers to a complete lack of transitions with rewards exceeding value threshold $\beta$, either due to a flawed policy $\pi$ (\textit{generative} and \textit{conceptual}), or ill-defined state and action spaces $S,A$ (\textit{hopeless}).

\subsection{Mapping \texorpdfstring{$\M_{\tau}$}{Mtau}}

\noindent\textbf{Universe \& Sub-universe.}
The \textit{universe} $\U$ includes  finite trajectories $\tau = (s,a,s',\dots,s_\text{last})$ across all conceivable state and action spaces. As in $\M_{s,a,s'}$, the \textit{sub-universe} $\P$ comprises only trajectories from the mapped \ac{MDP}. 

\smallskip
\noindent\textbf{Acceptability \& Conceptual Space.}
We map \textit{acceptability} $\N$ to a conditional probability distribution $p_{\tau}$ over variable-length trajectories: $p_{\tau}(\tau) := P(\tau\vert s,\pi,T) = \prod_{s,a,s'\in \tau} \pi(a)T_{a}(s,s')$. Our \textit{conceptual space} $\C$ comprises trajectories with this probability larger than $\alpha$.

\smallskip
\noindent\textbf{Evaluation.}
We map \textit{evaluation} $\V$ to some normalisation $\overline{V}$ of the agent's estimated value of the final state in the trajectory, $V(s_\text{last})$ such that $\overline{V}: \textit{tuples}(S\! \times\! A) \to [0,1]$.

\smallskip
\noindent\textbf{Traversal Strategy.}
\textit{Traversal strategy} $\Q$ takes tuples of trajectories as input and outputs the same trajectories appended with one new action and state resulting from a one-step rollout of policy $\pi$. %
During traversal, the observation of each new state rules out some potential trajectories and changes the probabilities of others. Consequently, conceptual space $\C$ changes during traversal in accordance with $\alpha$.

\smallskip
\noindent\textbf{Transformations.}
\textit{$\N$-transformation} refers to a change to policy $\pi$ such that conceptual space $\C$ changes. When we change our policy, the set of possible trajectories can also change. %
\textit{$\Q$-transformation} maps to any change to policy $\pi$. 

\smallskip
\noindent\textbf{Aberration \& Uninspiration}
\textit{Aberration} maps to experiencing unlikely trajectories given the trajectory so far, policy $\pi$, and transition probabilities $T$. A trajectory can have a low likelihood for two reasons: our policy can assign low probabilities to certain actions, or certain transition probabilities might be low according to $T$. Different types of aberration are categorised similarly as in mapping $\M_{s,a,s'}$: \textit{perfect}, \textit{productive}, and \textit{pointless} aberration correspond to situations where all, only some, or none of the aberrant trajectories are valued by the normalised value function $\overline{V}$ past the value threshold $\beta$. \textit{Uninspiration} is mapped similarly: \textit{generative}, \textit{conceptual}, and \textit{hopeless} uninspiration match with no valued trajectory reachable by policy $\pi$, with probability larger than $\alpha$, or expressible with given state and action spaces $S,A$. Additionally, the value function $\overline{V}$ may be flawed.

\section{Quality of Mappings}
\label{sec:discussion}
Our mappings resulted from an iterative and intuitive process, gradually incorporating insights about how our mapping choices interplay and whether they retain compatibility with \citeauthor{boden:creative_mind}'s original theory, the \ac{CSF}, and \acp{MDP}. 
Based on this process, we formulated quality criteria to guide researchers' choice of mappings. Our formulation was retrospective, non-exhaustive, and independent of additional theory or related work.
We demonstrate below that each criterion can surface issues of the quality of a given mapping via application to $\M_s, \M_{s,a,s'}$, and $ \M_\tau$.

\smallskip
\noindent\textbf{Semantic Similarity to Boden’s Original Theory}\\
A mapping should preserve as much of the meaning established in \citeauthor{boden:creative_mind}'s theory, once cast to the elements and dynamics of \acp{MDP}. Due to space limitations, we reflect on the central notion of \textit{transformations} only. %
\citet[p.~6]{boden:creative_mind} describes \textit{($\N$-)transformations} as the \enquote{deepest cases of creativity.} However, in \bm{$\M_s$},
mapping the state space $S$ to the conceptual space $\C$ renders $\N$-transformations impossible, as the agent cannot modify $S$, contradicting \citeauthor{boden:creative_mind}'s theory. 
To support generality in the design of our mappings, we imposed only minimal requirements (a policy) on our agent, and in this case found no natural way to allow changes to $\C$. 
Relaxing this design choice, a small change to $\M_s$ could overcome the issue, mapping between the conceptual space and an agent's current \textit{model} of the state space \cite[e.g.,][]{konstantinos:adaptive_state_partitioning}, rather than the MDP's \enquote{objective} state space. 
In \bm{$\M_{s,a,s'}$} and \bm{$\M_\tau$}, \textit{$\N$-transformation} refers to a change in the policy that changes the conceptual space. 
Thus, a conceptual space is partly determined by the policy. This association is semantically similar to \citeauthor{boden:creative_mind}'s description of conceptual spaces as a \emph{style} of thinking: \enquote{actions might be totally mental} \citep[p.~275]{sutton1997significance}, and a policy reacts to the situation and past experiences to decide what to do next.

\smallskip
\noindent\textbf{Functional Similarity to the \ac{CSF}}\\
In the \ac{CSF}, acceptability $\N$ and evaluation $\V$ are parameters to traversal strategy $\Q$. This gives $\Q$ \enquote{awareness} of conceptual space $\C$ and enables decision-making based on evaluation $\V$. In \bm{$\M_s$}, acceptability $\N$ is formalised as a membership function $\mu$ to state space $S$. 
$\Q$, then, does not make use of $\N$, as all possible outputs will themselves be states, and therefore elements of the conceptual space.
In contrast, evaluation $\V$, mapped to value function $\hat{V}$, quite reasonably informs $\Q$, as policies are commonly inferred from value functions. A problem regarding $\N$ as a parameter to $\Q$ also exists in \bm{$\M_{s,a,s'}$} and \bm{$\M_\tau$}: acceptability $\N$ is rarely available \citep[p.~275]{sutton1997significance} to the agent, which often only implicitly \enquote{experiences} the transition probabilities $T$ while rolling out a policy. 
Another concern with $\M_{s,a,s'}$ and $\M_\tau$ is the dependency between \textit{$\N$}- and \textit{$\Q$-transformation}: the former forces the latter, excluding the possibility of changing our acceptability $\N$ without change to traversal strategy $\Q$. In the \ac{CSF}, these two are independent components. Even though the behaviour of $\Q$ will be impacted by a change in $\N$ or $\V$, the traversal strategy itself will not change.

\smallskip
\noindent\textbf{Compatibility Between Frameworks}\\
Rewards and values are arguably the most semantically aligned choice with the \ac{CSF} evaluation function. However, in contrast to the latter, they are not meant to be squashed into the unit interval; their potential to extend the real line allows them to flexibly represent both real-world quantities and challenges. We leave open \textit{how} to normalise these quantities, introducing undesirable ambiguity.
As another compatibility issue, in \bm{$\M_s$}, 
\emph{aberrations} are left unmapped due to policies operating strictly inside state spaces. 

\smallskip
\noindent\textbf{Applicability Across Diverse AI Systems}\\
In $\bm{\M_s}$, acceptability $\N$ is defined as the membership function $\mu$, outputting binary values. For some systems, a continuous $\mu$ may be more appropriate; e.g., fuzzy definitions of state space $S$ have been used \citep{busoniu:fuzzy_state_space}. Alternatively, acceptability $\N$ could be correspond to a probability of a state belonging to state space $S$, perhaps combined with a notion of confidence \citep{grace:creative_expectation}.

\section{Identifying Types of Creativity}
We aim to identify \textit{exploratory} and \textit{transformational creativity} in our mappings. 
For brevity, \enquote{sufficiently valued} concepts here mean concepts that pass value threshold $\beta$.

\smallskip
\noindent\textbf{Exploratory Creativity} (finding novel valued concepts)
\begin{description}
\item[$\M_s$]: finding novel states with sufficient estimated value.
\item[$\M_{s,a,s'}$]: finding novel transitions with sufficient reward.
\item[$\M_\tau$]: finding novel trajectories with sufficient estimated value for the final state.
\end{description}

\smallskip
\noindent\textbf{Transformational Creativity}\\ (finding $\N$ and $\Q$ that admit novel valued concepts)
\begin{description}
\item[$\M_s$]: \textit{$\N$-transformational creativity} (\textit{$\N$-TC}) requires redefining the membership function $\mu$ such that the state space $S$ admits new states with sufficient estimated value. \textit{$\Q$-transformational creativity} (\textit{$\Q$-TC}) requires updating policy $\pi$ such that new valued states are found.
\item[$\M_{s,a,s'}$]: \textit{$\N$-TC} here is updating policy $\pi$ such that new sufficiently rewarded transitions are found, but with the additional requirement that the set of transitions that pass probability $\alpha$, i.e.~the conceptual space $C$, changes. \textit{$\Q$-TC} works similarly but without said requirement.
\item[$\M_\tau$]: \textit{$\N$-TC} and \textit{$\Q$-TC} are the same as in mapping $\M_{s,a,s'}$ but instead of  transitions we have entire trajectories.
\end{description}

\section{Critical Assumptions}
This research required a very close reading and comparison of \citeauthor{boden:creative_mind}'s original account and \citeauthor{wiggins:cf_2019}'/\citeauthor{ritchie:csf}'s formalisations, highlighting several inconsistencies. %

\citet[][p.~4]{boden:creative_mind} defines a conceptual space as \enquote{any disciplined way of thinking that is familiar to (and valued by) a certain social group}, suggesting that all concepts in the space are considered by default valuable. %
\citeauthor{wiggins:cf_2019} and \citeauthor{ritchie:csf} in contrast assume that membership in the conceptual space is only conditional on the \textit{typicality} of the concept/artefact, and that value is evaluated separately through $\V$ \citep[e.g.,][p.~28]{wiggins:cf_2019}. We adopt this view for its useful distinction between typicality and value.

We also consider \citet[p.~28]{wiggins:cf_2019} and \citet[p.~44]{ritchie:csf} describing $\V$ as defining \enquote{value} problematic; it suggests the evaluation is \emph{only} of value, leaving open how the other core components of creativity, most importantly \textit{novelty} but also \textit{surprise} \citep[][p.~1]{boden:creative_mind}, are assessed. Here, we assume that $\V$ may or may not evaluate these core components. In neither mapping do we specify which components the reward or value function mapped to evaluation capture; consequently, we leave it open whether traversal under this mapping actively promotes exploratory creativity.

\section{Related Work}
We relate our contributions to Colin et al.'s \citeyearpar{colin:hierarchical_mdp_csf} work as the only close predecessor in terms of goals and methods. They mapped the \ac{CSF} as formalised by \citet{ritchie:csf} to hierarchical \acp{MDP} as instantiated in the options framework proposed by \citet{sutton1999between}. Their object-level conceptual space comprises policies, evaluation is mapped to a discounted return function %
and traversal is mapped to a policy update function. Via online learning (e.g. standard temporal difference learning), an agent traverses the space of possible policies. On the meta-level, it traverses pairs of discounted return and policy update functions%
, with meta-level evaluation assessing the value of a given pair for solving the problem at hand. The policy update function is constrained to evolving policies within a single option only, i.e.~policies that start from a specific set of initiation states, and end on a specific termination condition. Traversal over these pairs changes which option and underlying space of potential policies is used to tackle the problem at hand. 

Colin et al.'s object-level thus matches our proposed mapping $\M_\pi$, which we did not pursue further as we considered it a meta-level interpretation. %
Crucially though, the \ac{CSF} allows for many meta-levels and Colin et al. thus complement the present work by contributing a candidate mechanism to facilitate transformational creativity on the space of policies by extending MDPs to include behavioural hierarchies. Crucially though, the focus on hierarchical \acp{MDP} and the options framework as only one (albeit popular) formalisation thereof limits the wider applicability of their findings. 
We, in contrast, map \acp{MDP} as a more general but less expressive framework to increase the applicability of our findings while understanding \acp{MDP}' limitations w.r.t.\ modelling creativity. %

As a second distinction, Colin et al. propose only a single mapping, not motivated beyond its relevance to robot control. In contrast, we discuss three of eleven potential mappings in depth, complemented with domain-agnostic quality criteria for probing and comparing different mappings. %

As a third and final distinction, Colin et al. focus on computational models of insight, corresponding to transformational creativity, but do not integrate uninspiration as threats to, and aberration as opportunities for creativity. Related, they do not consider $\N$ part of the explored meta-level triplets, hence providing a mapping for $\Q$-transformation and revisions to $\V$ only. Despite being interested in insight, they thus do not explicitly address $\N$-transformation as the most widely known and accepted type of transformational creativity, and only instance presented by \citet{boden:creative_mind} originally. %
We interpret aberration, uninspiration and transformational creativity under each mapping, thus embracing \citeauthor{boden:creative_mind}'s theory and the \ac{CSF} more comprehensively.

\section{Conclusion \& Future Work}
We have set out to provide theoretically grounded tools for the assessment of creativity in \ac{AI} systems at large. To this end, we put forward eleven, and discuss in detail three, formal mappings between agents in interaction with \acfp{MDP} and Margaret \citeauthor{boden:creative_mind}'s process theory of creativity, using the \acf{CSF} as a stepping stone. We leveraged these three mappings to reflect on the types of, opportunities for, and threats to creativity conceivable in a system formalised on an \ac{MDP}. 

Our findings motivate exploration of further mappings as the most imminent future work, which will be supported by our critical reflection on quality criteria, and our discussion on the  constraints on potential mappings imposed by formal features of \acp{MDP} and the \ac{CSF}. The influence of hyperparameters $\alpha$ and $\beta$ must also be investigated. This object-level effort should be complemented with formalising the corresponding meta-levels to highlight not only which forms of creative transformations are possible, but also how they can be brought about. This can for instance shed more light on the mechanisms behind policy changes and, in consequence, different forms of transformational creativity. Following this exploratory research, the best mapping candidates should be further tested through application to existing systems that have been attributed creativity, or that realised major AI milestones for which creativity is commonly considered necessary. 
Our present focus on \acp{MDP} allowed us to define potential mappings by trading off simplicity and widespread applicability; as another avenue for future work and prerequisite for evaluating the creativity of more constrained but realistic systems, e.g., with partial observability and a need to learn models of the world, we recommend leveraging this foundation to integrate creativity theory and more general--but also more complex--sequential decision-making frameworks such as \acp{POMDP}.%

The synthesis of established creativity theory and AI frameworks can significantly enhance our understanding of, and consequently the potential for, creativity in \ac{AI}. We invite researchers from Psychology and \ac{AI} more generally and \ac{CC} specifically to join this interdisciplinary effort.

\section{Acknowledgements}
We thank Geraint Wiggins for helpful clarifications and our anonymous reviewers for their comments and encouragement for the vision presented here. NMA was supported by a Helsinki Institute for Information Technology fellowship.

\bibliographystyle{iccc}
\bibliography{iccc}

\end{document}